\documentclass[3p, twocolumn]{elsarticle}

\usepackage{lineno,hyperref}
\usepackage{subfigure}
\usepackage{caption}
\usepackage{float}
\usepackage{todonotes}
\usepackage{xcolor, colortbl}
\usepackage{array}
\newcolumntype{M}[1]{>{\centering\arraybackslash}m{#1}}
\definecolor{mygreen}{RGB}{174, 234, 174}
\definecolor{myred}{RGB}{255, 153, 128}

\modulolinenumbers[5]

\begin{document}

\begin{frontmatter}

\title{Knowledge Triggering, Extraction and Storage via Human-Robot Verbal Interaction}

\author[mymainaddress]{Lucrezia Grassi}\corref{mycorrespondingauthor}\cortext[mycorrespondingauthor]{Corresponding author}
\ead{lucrezia.grassi@edu.unige.it}

\author[mymainaddress]{Carmine Tommaso Recchiuto}

\author[mymainaddress]{Antonio Sgorbissa}

\address[mymainaddress]{Department of Informatics, Bioengineering, Robotics and Systems Engineering, University of Genoa, Via all’Opera Pia 13, 16145 Genoa, Italy}

\begin{abstract}
This article describes a novel approach to expand in run-time the knowledge base of an Artificial Conversational Agent. A technique for automatic knowledge extraction from the user's sentence and four methods to insert the new acquired concepts in the knowledge base have been developed and integrated into a system that has already been tested for knowledge-based conversation between a social humanoid robot and residents of care homes. The run-time addition of new knowledge allows overcoming some limitations that affect most robots and chatbots: the incapability of engaging the user for a long time due to the restricted number of conversation topics. The insertion in the knowledge base of new concepts recognized in the user's sentence is expected to result in a wider range of topics that can be covered during an interaction, making the conversation less repetitive. Two experiments are presented to assess the performance of the knowledge extraction technique, and the efficiency of the developed insertion methods when adding several concepts in the Ontology.
\end{abstract}


\end{frontmatter}


\section{Introduction}
\label{sec:introduction}
``I want to talk about my \textit{wife}..." ``If you don't know what to choose, I can show you some of the things I can do on my tablet." ``No...I want to talk about my \textit{wife}...she was only 71 when she died." ``I see, very interesting. I am aware that traditionally English people were either Protestants or Catholic." ``I want to talk..." ``If you don't mind I will ask you a question..."

This very frustrating conversation between a care home resident and a fully autonomous Pepper robot happened on $7/18/2019$ in the UK, during an experimental trial that, generally speaking, proved to be successful under many perspectives \cite{Papadopoulos2021}. During the conversation, the robot was not capable to correctly interpret the intention of the person to talk about his wife because of a bug, which was fixed immediately after the session. The day after, the conversation with the user went like this: ``I'll be glad to talk about my \textit{wife} and my children." ``Just to confirm: can we talk about your \textit{wife}?" ``Yes." ``Your \textit{wife} must be a very nice person." ``Yes..." ``Tell me more about your wife, if you like". After counter-checking that the man really wanted to talk about his wife, considered a sensitive issue, a long conversation started giving the man the opportunity to recall and share memories about his dead wife.
The second, and more successful, interaction was possible because, after fixing the bug, the robot could rely on a huge knowledge base designed, by hand, by knowledge engineers supported by nursing experts, which included many topics of conversation related to daily life. But what if the concept of ``wife", possibly very relevant for an older man living in a care home, had not been inserted at all in such a knowledge base?

\begin{figure}
    \centering
    \includegraphics[width=\linewidth]{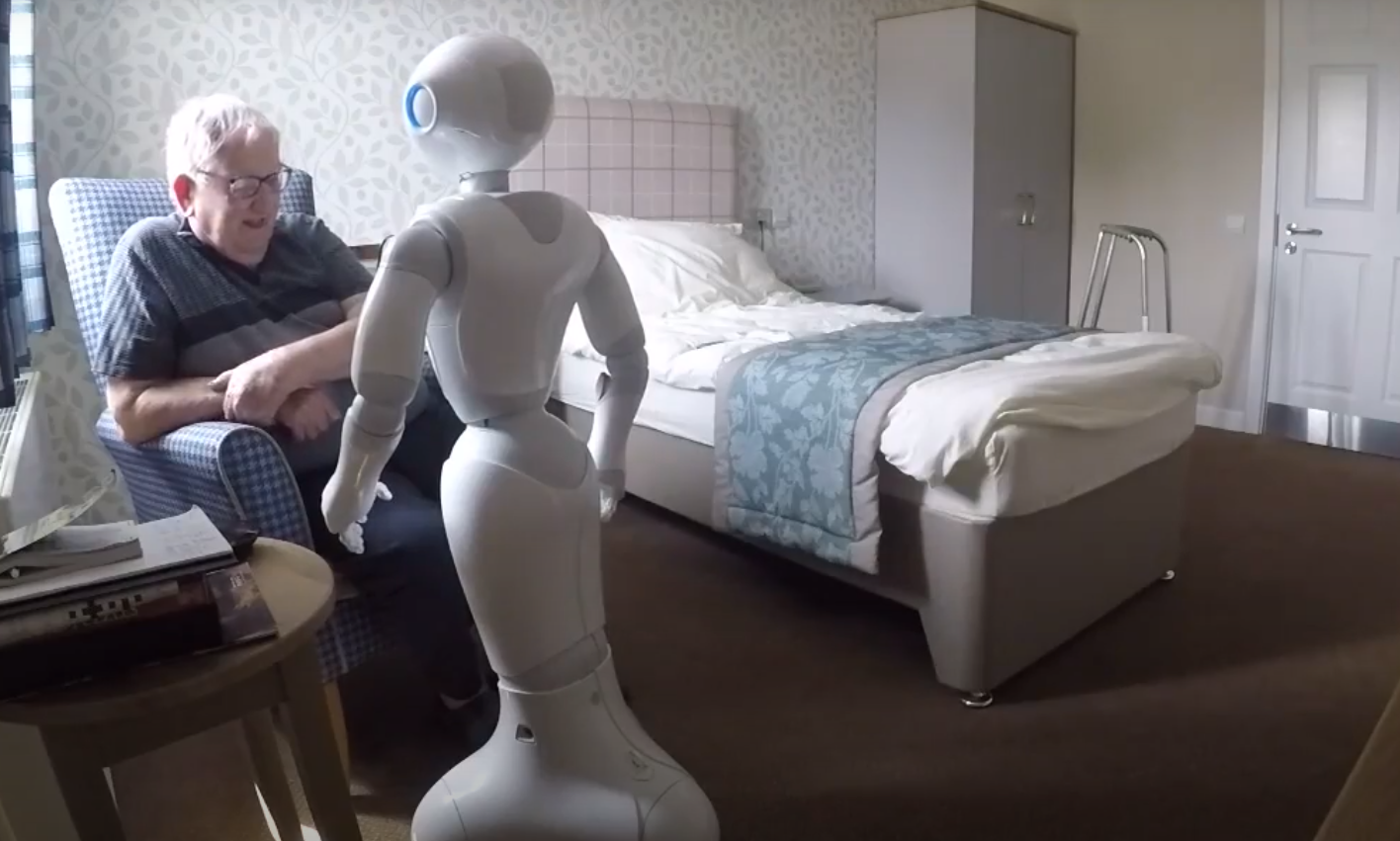}
    \caption{Autonomous conversation with Pepper during the CARESSES trial.}\label{fig:architecture}
\end{figure}   

Social Robotics is a research field aimed at providing robots with a brand new set of skills, specifically related to social behaviour and natural interaction with humans.
Social robots can be used in many contexts such as education \cite{Belpaeme2018}, stress management \cite{Yorita2018}, welcoming guests in hotels \cite{Pan2015}, cruise ships \cite{Pandey2018}, malls \cite{Niemela2019}, and elderly care \cite{Broekens2009}. A noteworthy application of Socially Assistive Robots (SARs) is in the healthcare field: it has been argued that robots can be used to improve mental health and make people feel less lonely and help to relieve caregivers' burden in care homes and domestic environments \cite{Papadopoulos2020}.

In this scenario, the recent EU-Japan project CARESSES\footnote{\href{http://caressesrobot.org/}{http://caressesrobot.org/}} is the first project having the goal of designing SARs that are culturally competent, i.e., able to understand and talk about the cultural beliefs, values, habits, customs, and etiquette of the person they are assisting, while autonomously reconfiguring their way of acting and speaking \cite{Bruno2017}. 
In the context of the project, a framework for autonomous, culturally competent conversation has been developed \cite{Sgorbissa2018}, \cite{Bruno2019}. The conversation framework can achieve mixed-initiative dialogues by exploiting the hierarchical structure of a Cultural Knowledge Base implemented as an OWL2 Ontology, thus enabling rich, knowledge-grounded conversations \cite{Recchiuto2020UR}. Being paralleled by a framework for probabilistic reasoning, such Ontology is designed to take into account the possible cultural differences between different users in a non-stereotyped way, and it stores chunks of sentences that can be composed in run-time, therefore enabling the system to talk about the aforementioned concepts in a culture-aware and engaging way \cite{Recchiuto2020RAL}\cite{Bruno2019IJSR}.

However, when trialing such a system with recruited participants, it is straightforward to observe that an Ontology manually crafted by knowledge engineers and nursing experts has some limitations: to avoid undesirable situations such as the one presented at the beginning of the article, the overall conversational capabilities of the system should allow the knowledge-base to be expanded in run-time.  

The article focuses on the problem of the acquisition of new knowledge during the conversation with the user, and the insertion of the recognized concepts in the knowledge base to make the conversation richer, more natural, and engaging.

The main contributions of the article are:
\begin{enumerate}
    \item A solution to recognize meaningful concepts from the users' sentences that can be promising candidates for insertion. 
    \item A portfolio of insertion solutions allowing the user to expand the knowledge base in run-time by enriching it with the recognized concepts.
\end{enumerate}  

The developed solutions are tested by involving (1) Amazon Mechanical Turk workers for the collection of sentences that shall trigger the extraction of new knowledge, as well as (2) a professional English translator to interactively insert the extracted knowledge in the Ontology through verbal interaction with the system. It is worth observing that the present article tests the aforementioned solutions only from a technical perspective, by providing statistics in terms of (1) true/false positive/negative rates in extracting concepts and (2) the average number of interaction steps required for insertion with different algorithms, whereas it completely ignores the subjective experience of users, to be evaluated through a proper protocol involving recruited participants as future work.

The remainder of this article is organized as follows. Section \ref{sec:SOTA} presents an overview of previous works related to techniques for knowledge automatic extraction and knowledge crowdsourcing; moreover, it gives insights into the CARESSES conversational framework as a starting point for the following discussion. Section \ref{sec:methodology} describes the approach for the recognition of concepts and the insertion methods. Section \ref{sec:experiments} describes the two experiments carried out: the first one aims to assess the performance of the method used to recognize and extract concepts from the user's sentence, while the second one is meant to compare the performance of the knowledge insertion methods. Section \ref{sec:results} presents and discusses the results obtained. Eventually, Section \ref{sec:conclusion} draws the conclusions.

\section{State of the Art}
\label{sec:SOTA}
This Section addresses techniques for knowledge automatic extraction and knowledge crowdsourcing, presenting a brief analysis of the related Literature. Furthermore, it gives an overview of the way in which the Ontology is structured in the CARESSES framework, it describes how the Dialogue Tree is generated starting from it and the related Dialogue Management Algorithm.

\subsection{Techniques and tools for knowledge automatic extraction}
Approaches for knowledge automatic extraction are typically based on the Ontology learning ``layer cake" \cite{Cimiano2006}, which foresees a stack of subtasks: extracting terms, synonyms, concepts, concept hierarchies, relations, relation hierarchies, axioms schemata, and general axioms \cite{Asim2018}. Two main classes of approaches have been investigated for implementing these tasks: \textit{Linguistic techniques}, which play a key role in all layers of the learning cake; \textit{Statistical methods}, mostly used for terms, concepts, and relations extraction. 

Linguistic techniques play a crucial role in pre-processing text-corpora through speech tagging, parsing, and lemmatization \cite{Petit2017}, and are commonly used to extract terms, concepts, and relations by analyzing the syntactic structure of phrases. For instance, this can be done through word sub-categorization combined with clustering techniques \cite{Faure1999}; using domain-specific seed words \cite{Fraga2017} that provide a base to extract similar domain-specific terms and concepts; using information deriving from parsing trees to find relationship patterns between terms \cite{Sordo2015}. Other linguistic methods commonly used for knowledge extraction consist in extracting terms and concepts using verb subcategorization \cite{Gamallo}, lexico-syntactic patterns by defining regular expressions \cite{Atapattu}, semantic lexicons \cite{Navigli}, or Convolutional Neural Networks \cite{waldis}.

Statistical techniques include C/NC value, a multi-word terminology extraction method \cite{Frantzi2000} that has been used to detect candidate concepts from different domains \cite{Chandu2017}, as well as co-occurrence analysis, a concept extraction technique that locates lexical units that usually occur together to find associations between different terms, e.g., using Chi-Square \cite{Frikh2011}, Mutual Information \cite{Xiao2016}, K-means Clustering or Latent Semantic Analysis (LSA) \cite{Rani2017}. Also, hierarchical clustering has been used to the aim of extracting taxonomic relations among terms, by using agglomerative clustering (bottom-up approach) \cite{zepeda} or divisive clustering (top-down approach) \cite{dhillon}. Finally, Association Rule Mining (ARM) has been used to find hidden non-taxonomic relations and patterns, by adopting A-priori algorithms \cite{idoudi} and FP growth algorithms \cite{paiva}.

Whatever approach is adopted, Literature surveys conclude that automatic Ontology learning is more effective when relying on seed words or a base Ontology instead of building it from scratch \cite{Asim2018}. In line with this principle, also our approach relies on the assumption that an initial Ontology manually designed by experts exists, possibly re-using existing Ontologies \cite{Chen2014}. 
However, as better specified in the following, the system described in this work will rather propose a semi-automatic extraction and insertion of new knowledge, through an explicit interaction with the user. The reason for giving priority to a supervised approach is straightforward: even if the data automatically extracted (e.g., from the Internet) are subjected to quality control to remove personally identifiable information, messy code, inappropriate content, spelling mistakes, etc., a fully automated system not supervised by humans may not be appropriate in many situations (e.g., in care homes with old people or with children) that are typically considered as a top-priority area for SAR applications.

\subsection{Knowledge crowdsourcing through knowledge-grounded conversation}
The possibility of acquiring knowledge by relying on networked interactions with people has been recently explored in different domains. For example, some museum collections have used a ``social tagging" approach to enhance curatorial documentation \cite{Trant2009}, and crowdsourcing mechanisms for collaborative Ontology construction have been investigated as well \cite{Zhitomirsky2017}. 

Recently, crowdsourcing approaches have been adopted for supporting conversational tasks of social robots, in situations when the robot needed to act as a life coach \cite{Abbas2020}, or assist end-users with information retrieval \cite{Lasecki2013}. However, in these contexts crowd workers recruited in real-time were directly teleoperating the robot’s speech, without actually increasing the robot's knowledge base.

To the best of our knowledge, the only example of knowledge crowdsourcing for verbal interaction is Alexa Answers, an online platform where Alexa users may provide answers for questions that were previously unanswered by the device. The platform raised some criticisms: in particular, because of a lack of control of the answers’ quality, which is sometimes untrue, potentially sponsored, and offensive, notwithstanding a peer rating points-based system \cite{Wiggers2021}. Moreover, customers' answers are only used for reactively replying to a specific question by other users, and the acquisition of new knowledge does not happen in real-time through verbal interaction with Alexa, but it is implemented as a separate process through a web form. 

Some popular bots such as XiaoIce \cite{Zhou2020} adopt a hybrid approach since they have a topic database that is periodically updated by collecting data from two data sources: human conversational data from the Internet, (e.g., social networks,
public forums, news comments, etc.), and human-machine conversations generated by XiaoIce and her users. However, the process is aimed at collecting query-response pairs from a human-machine conversation and does not allow the user to explicitly add a new topic in the topic database or sentences to talk about such a concept.

The knowledge-grounded chit-chatting framework described in \cite{Mavridis2015} allows users to enrich the knowledge base by acquiring new knowledge about them and their personal preferences to avoid stereotyped representations of people and cultures. However, it only allows the user to express his/her attitude towards a given concept: functionalities to add new concepts and sentences to  appropriately talk about such concepts are completely absent.

As the reader can observe, and also in virtue of the challenges it poses from an ethical perspective, the number of conversational systems in the Literature allowing the user to customize the interaction by adding new concepts and sentences is very small. Ideally, the following desiderata shall be met: engaging the users to talk freely (e.g., about their memories, preferences, plans, etc.) to the end of triggering detection of concepts that are not yet encoded in the Ontology; introduction of the new concept in the Ontology hierarchy under the guidance of the person through taxonomy-based interaction [\cite{Bruno2018}, \cite{Bruno2019}]; creation of new instances in the Ontology encoding the attitude and considerations of the person towards a given concept, to produce a richer and more engaging interaction with the same person in the future or with other people using the Cloud (other people will then be able to express their attitude or add their sentences by adding new instances of the same concepts). However, to the best of our knowledge, there are no systems currently addressing all these desiderata.

Finally, it shall be mentioned that the problem of moderating crowdsourcing through administrators has been addressed \cite{Sher2020}, which is key to ensure that only appropriate concepts and sentences are added when different users, possibly with different sensitivities, contribute to enriching the knowledge base. Aggregation measures and algorithms to find consensus among multiple users \cite{Zhitomirsky2016}, as well as peer rating mechanisms \cite{Wagenknecht2017}, have been explored as well.

\subsection{Ontology and Dialogue Tree}
\label{sec:Ontology and DT}
Knowledge-based conversational agents and smartphone applications have been widely investigated in recent years, with a particular focus on the healthcare domain \cite{Jung2016}. In the conversational system taken as a reference in this work \citep{Recchiuto20206559}, the ability of the companion robot to naturally converse with the user has been achieved by creating a framework for cultural knowledge representation that relies on an Ontology \cite{Guarino1998} implemented in OWL2 \cite{Motik2008}. According to the Description Logics formalism, concepts (i.e., topics of conversation the system is capable of talking about) and their mutual relations are stored in the terminological box (TBox) of the Ontology. Instead, instances of concepts and their associated data (e.g., chunks of sentences automatically composed to enable the system to talk about the corresponding topics) are stored in the assertional box (ABox). 

To deal with representations of the world that may vary across different cultures \cite{Carrithers2010}, the Ontology is organized into three layers and explained more in detail in \cite{Recchiuto2020RAL}\cite{Bruno2019IJSR}. The TBox (Figure \ref{fig:architecture}) encodes concepts at a generic, culture-agnostic level and it should include concepts that are typical of the union all cultures considered, whichever the cultural identity of the user is, to avoid stereotypes. An example related to different kinds of beverages is shown: the system will initially guess the user's preferred beverages, but it will then be open to considering choices that may be less likely in a given culture, as the user explicitly declares his/her attitude towards them. For instance, the system may initially infer that an English person may be more interested to talk about \texttt{Tea} rather than \texttt{Coffee}, and the opposite may be initially inferred for an Italian user. However, during the conversation, initial assumptions may be revised, thus finally leading to a fully personalized representation of the user's attitude towards all concepts in the TBox, to be used for conversation.


\begin{figure}
    \centering
    \includegraphics[width=\linewidth]{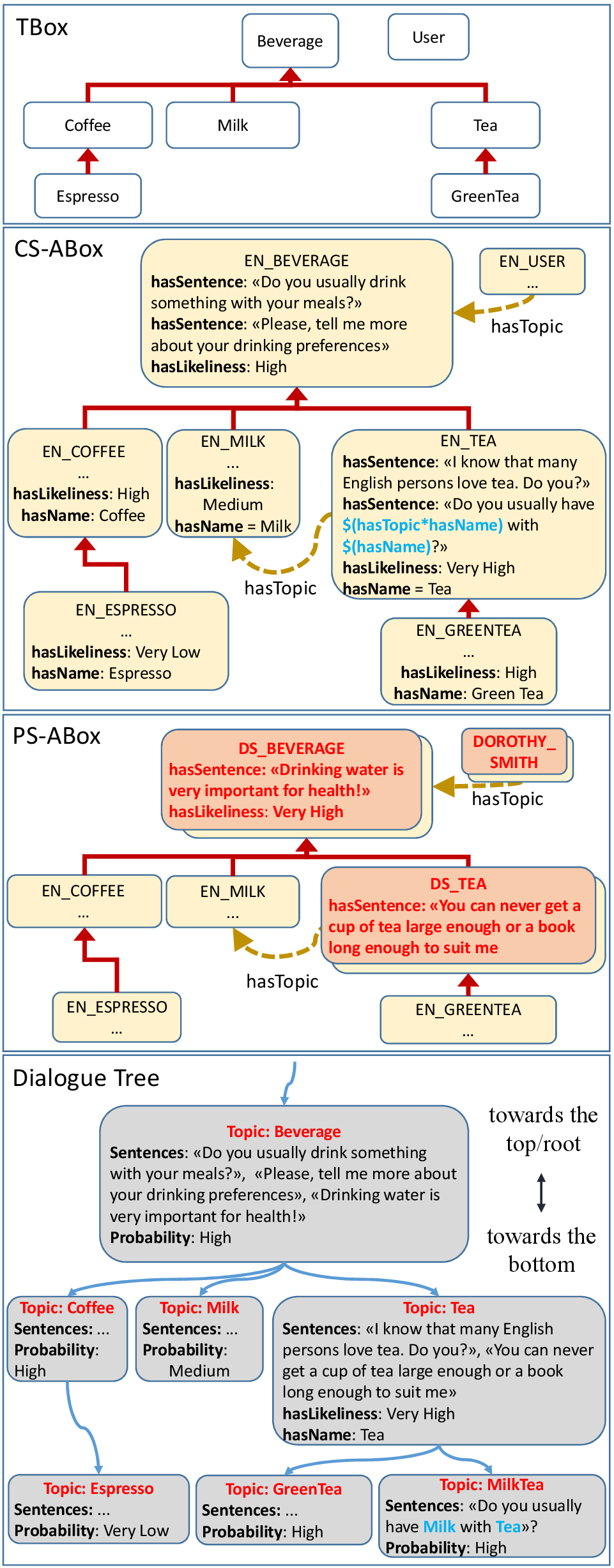}
    \caption{The three layers of the Ontology: TBox, CS-ABox (for the English culture), PS-ABox (for Dorothy Smith), and the Dialogue Tree generated from the Ontology structure.}
    \label{fig:three-layers}
\end{figure}

To implement this mechanism, the Culture-Specific ABox layer (CS-ABox in Figure \ref{fig:architecture}) comprises instances of concepts encoding culturally appropriate chunks of sentences to be automatically composed (Data Property \texttt{hasSentence}) and the probability that the user would have a positive attitude toward that concept, given that he/she belongs to that cultural group (Data Property \texttt{hasLikeliness}). 
Eventually, the Person-Specific ABox (PS-ABox in Figure \ref{fig:architecture}) comprises instances encoding the actual user’s attitude towards a concept updated during the interaction (the system may discover that Mrs Dorothy Smith is more familiar with having tea than the average English person, instance \texttt{DS\_TEA} with \texttt{hasLikeliness=Very High}), sentences to talk about a topic explicitly encoded/added to the system by the caregivers (\texttt{hasSentence}=``You can never get a cup of tea large enough or a book long enough to suit me") or other knowledge about the user (e.g., name, the town of residence) explicitly added during setup. 
At the first encounter between the robot and a user, many instances of the Ontology will not contain Person-Specific knowledge: the robot will acquire awareness about the user's attitude at run-time, either from its perceptual system or through verbal interaction, e.g., asking questions. 
For a detailed description of the terminological box (TBox) and assertional box (ABox) of the Ontology, as well as the algorithms for cultural adaptation, see \cite{Bruno2019}.

The Dialogue Tree (DT) (Figure \ref{fig:architecture}), used by the conversation system to chit-chat with the user, is built starting from the Ontology structure: each concept of the TBox and the corresponding instances of the ABox are mapped into a conversation topic, i.e., a node of the tree; the Object Property \texttt{hasTopic} and the hierarchical relationships among concepts and instances are analyzed to define the branches of the DT. In the example of Figure \ref{fig:architecture}, the instance of \texttt{Tea} for the English culture is connected in the DT to its child node \texttt{GreenTea} (which is a subclass of \texttt{Tea} in the Ontology), and its sibling \texttt{MilkTea} (since \texttt{EN\_MILK} is a filler of \texttt{EN\_TEA} for the Object Property \texttt{hasTopic}). 

Each conversation topic has chunks of culturally appropriate sentences associated with it, that are automatically composed and used during the conversation. Such sentences can be of different types (i.e., positive assertions, negative assertions, or different kinds of questions) and may contain variables that are instantiated when creating the tree. For instance, a hypothetical sentence ``Do you like \$hasName?" encoded in the concept \texttt{Coffee} might be used to automatically produce both ``Do you like Coffee?" and ``Do you like Espresso?" (being \texttt{Espresso} a subclass of \texttt{Coffee}). 
In the base version used for testing, exploiting the taxonomy of the Ontology allowed us to easily produce a DT with about 3.000 topics of conversation and more than 20.000 sentences, with random variations made in run-time.

Based on the Dialogue Tree, the key ideas for knowledge-driven conversation can be briefly summarized as follows (the whole process \cite{Recchiuto2020RAL} is obviously more complex). 

Each time a user sentence is acquired:
\begin{enumerate}
    \item A Dialogue Management algorithm (either keyword-based or based on more advanced topic classification techniques) is applied to check if the user's sentence may trigger one of the topics in the DT by jumping to the corresponding node;
    \item If no topics are triggered, the conversation
    follows one of the branches of the DT (according to policies that take into account the user's cultural background and personal preferences).
\end{enumerate}
    
The system continues in this way, proposing sentences corresponding to a node and acquiring the user's feedback that can be used to update the user's preferences and/or determine the next node to move to.

However, in the original version of the system, if a concept or topic of conversation raised by the user is not encoded in the Ontology, there is no way for the user to explicitly add it in run-time during verbal interaction, nor to add corresponding sentences to be encoded in the Ontology as Class restrictions and Data Properties, thus limiting the system's versatility and the user's experience.

\section{Methodology}
\label{sec:methodology}
Most systems (including the one described in the previous section), whenever they do not recognize the user's intent, activate a standard default behaviour that, in the worst case, may lead to the frustrating reply ``Sorry, I do not understand". Even when the system correctly understands what the user says, the risk is that the same conversation topics are repeatedly addressed, and the dialogue becomes less engaging. To overcome these limitations, it is important to be able to expand in run-time the knowledge base with new topics of conversation raised by the user as well as new sentences to talk about those topics. 

The overall behaviour of the Dialogue Management algorithm is the same as described in the previous Section: however, if the user's sentence does not trigger any topic (case 2), the following alternative behaviour is implemented:

\begin{enumerate}
\setcounter{enumi}{1}
    \item If no topics are triggered:
    \begin{enumerate}
    	\item The user's sentence is analyzed to check if an opportunity to extract a new concept that is relevant for the user arises, possibly using third-party systems for Natural Language Processing (Section \ref{sec:methodology-recognition}).
    	\item If something relevant is returned, a method for the insertion of the new concept in the TBox of the Ontology is called (Section \ref{sec:methodology-insertion});
    	\item The user is asked to formulate a sentence that may be later used by the system to talk about that concept; the sentence is stored in the ABox of the Ontology and possibly re-used with other users of the system as well after being approved by a moderator (not discussed in this article).
    \end{enumerate}
\end{enumerate}

Steps 2.a and 2.b are discussed in greater detail. 

\subsection{Recognition of concepts}
\label{sec:methodology-recognition}
In the first step, a mechanism to recognize relevant concepts mentioned by the user during the conversation is required.

To this purpose, a Dialogflow Agent\footnote{\href{https://dialogflow.cloud.google.com/}{https://dialogflow.cloud.google.com/}}, i.e., a web service to manage functionalities for autonomous conversation, has been created. Dialogflow Agents are characterized by \textit{Intents}, which aim to capture users' intention for one conversation turn, and \textit{parameters}, which are values extracted from the sentence depending on the training of the Agent. 

In the current version of the system, four Intents have been defined: \textit{memories-past}, \textit{preferences}, \textit{norms}, and \textit{beliefs}, corresponding to broad areas of conversation (i.e., high level topics) already present in the starting Ontology. These specific Intents have been chosen because of their relevance in the development of SARs for taking care of older people: indeed, guidelines prepared by Nursing experts as well as experiments performed with older people in care homes \citep{Papadopoulos2020, Papadopoulos2021} clearly show the importance to engage older people to talk about their memories, what they like or dislike, what they believe to be important (something that can radically vary in different cultures). According to this, the Dialogue Management System is programmed in such a way that, from time to time, it engages the user to talk freely to trigger a reply that may contain useful knowledge to be extracted.

Dialogflow Intents need to be trained with a set of example phrases of what users might say, which may include manually tagged parameters. As an example, the \textit{preferences} Intent may contain training phrases such as ``I love $<$tagged parameter$>$" or ``I like $<$tagged parameter$>$": whenever the user says ``I love $<$concept$>$" or a variant of the same sentence such as ``Man, I really enjoy $<$concept$>$ for dinner" the sentence is matched to the \textit{preferences} Intent and whatever the value of \textit{concept} is, it is returned as a response. Notice that Dialogflow, if properly trained, allows the designer to tag more than one word as belonging to the same parameter, and therefore the same Intent may be capable to recognize ``playing baseball" or ``going to school" as something that the user likes or dislikes. 

To collect appropriate training phrases for each Intent in such a way that Dialogflow can correctly perform Intent matching, a vocal questionnaire has been created. To have a reasonable amount of sentences, we recruited 30 workers through Amazon Mechanical Turk\footnote{\href{https://www.mturk.com/}{https://www.mturk.com/}} (MTurk). 
Workers were required to answer, using their microphone, 20 questions (5 for each Intent):

\begin{enumerate}
    \item Memories/Past
    \begin{enumerate} 
        \item \textit{How have things changed compared to when you were young?}
        \item \textit{Please tell me about your childhood.}
        \item \textit{Please tell me about your childhood friends.}
        \item \textit{What games did you use to play when you were a child?}
        \item \textit {Please tell me about the school you went to.}
    \end{enumerate}
    \item Preferences
    \begin{enumerate} 
        \item \textit{How do you like to spend time with your friends?}
        \item \textit{What are your favourite foods?}
        \item \textit{Please tell me about your favourite songs.}
        \item \textit{I would love to know about the movies you like and your favourite actors.}
        \item \textit {What is your favourite animal?}
    \end{enumerate}
    \item Norms
    \begin{enumerate}
        \item \textit{Please tell me how to celebrate the birthday of a loved one.}
        \item \textit{As your robot assistant, how should I behave with your friends?}
        \item \textit{Please tell me about the good manners that matter to you.}
        \item \textit{What are the things people shouldn't do in the presence of others?}
        \item \textit {Are there any foods or drinks people should avoid?}
    \end{enumerate}
    \item Beliefs
    \begin{enumerate}
        \item \textit{What do you think about life?}
        \item \textit{I would like to know what you think about religion.}
        \item \textit{I would like to know what you think about marriage.}
        \item \textit{Do you think family relationships are important? Please tell me your thoughts.}
        \item \textit {People say: ``Healthy body in healthy mind". I would be happy to know what you think about this.}
    \end{enumerate}
\end{enumerate}

The answers of the first 20 submitted questionnaires have been used for the training, while the remaining 10 questionnaires have been used to test the mechanism (see Section \ref{sec:experiment-DF}).

It shall be stressed that thanks to how Dialogflow works, the answers to the aforementioned questions are not meant to train the system to understand specific sentences about a specific topic. For instance, a sample reply to the question  ``What is your favourite animal?" is ``My favourite \textit{animal} is the \textit{dog}", but the same syntactic structure can be used to train the system to understand any generic sentence like ``My favourite $<$concept$>$ is the $<$concept$>$". Similarly, a sample reply to the question ``Please tell me about the school you went to" is ``I went to a \textit{public school} near \textit{my house} and it was a pretty good \textit{school}", whose syntactic structure can be used to train the system to understand specific sentences such as ``I went to a \textit{swimming pool} near the \textit{school} and it was a pretty good \textit{swimming pool}". Summarizing, the questionnaire addresses broad areas of conversation that are expected to trigger prototypical replies to train the system, but the specific concepts that will be later extracted during run-time interaction are not uniquely determined by such replies.

In run-time, when the user is actually interacting with the system, the sentence is sent both to the pre-trained Dialogflow Agent for Intent recognition and to Cloud Natural Language\footnote{\href{https://cloud.google.com/natural-language/docs}{https://cloud.google.com/natural-language/docs}} (CNL), another Google Web service performing basic NLP operations.
Depending on the training sentences used, Dialogflow returns a candidate concept for the insertion (if any) for every piece of the user's sentence delimited by the ``and" stop word, while CNL takes the whole user's sentence and returns all the nouns (i.e., Entities) recognized in it as well as their associated Entity Type (e.g., \texttt{Person}, \texttt{Location}, \ldots, \texttt{Other}, etc.). To be candidates for insertions, concepts returned by Dialogflow (e.g., \textit{dog} or \textit{swimming pool} in the previous examples) need to include an Entity recognized by CNL: this is a safety measure to prevent that Dialogflow extracts a standalone adjective or adverb. Then, all candidates for insertions are ordered based on the priorities we assigned to Entity Types, e.g., concepts including \texttt{Person} is given a higher priority than concepts including \texttt{Location}. The extracted concept with the highest priority is chosen as the best candidate for the insertion. Eventually, such a concept is lemmatized, before proceeding with its insertion in the Ontology.

\subsection{Insertion of concepts}
\label{sec:methodology-insertion}
The Entity Type returned by CNL is not sufficient for knowing the exact position of the new concept to be added in the taxonomy: even if its Entity Type matches with one of the classes of the Ontology, there may exist subclasses which are more indicated as parent class of the new concept. Thus, once a concept has been extracted from the user's sentence, it is essential to develop an efficient strategy allowing the user to add this concept in the most appropriate place of the Ontology. 

Specifically, four methods are developed for the insertion of new concepts: (i) a Depth First Method, (ii) a method that exploits the Entity Type associated with the extracted concept, (iii) a method based on a definition of the concept provided by the user, and (iv) a method based on a generic sentence about that concept (not necessarily a definition), once again provided by the user. Depending on the extracted concept, one strategy could be more efficient than another: the method used for the insertion can be easily changed each time the algorithm is called.

In the following subsections, the four insertion methods are described.

\subsubsection{Depth First Method}
\label{brute-force}
To insert a new concept such as ``Espresso coffee", the user is asked a series of ``yes/no" questions, starting from the root of the DT (below the top concept \textit{Topic}), and following the taxonomy defined by the Ontology. That is, given that ``Coffee" is a subclass of ``Beverage" (Figure \ref{fig:architecture}), the algorithm will first ask ``Is espresso coffee a type of beverage" and then move to ``Is espresso coffee a type of coffee".
        
Suppose now that the concept ``Orange juice" needs to be added. At any given level, the system will ask multiple ``yes/no" question to choose the subclass/path to follow to a lower level of the tree (e.g., the class ``Beverage" may have the children ``Coffee", ``Milk" and ``Tea"): if the user answer ``no" to all questions and there are no more children in that level, he/she is asked if the concept can be inserted as a new child of the parent class (i.e., a new sibling of ``Coffee", ``Milk" and ``Tea"). If the answer is positive, the new concept is added to the Ontology: otherwise, the insertion procedure terminates and the concept is not added. The same happens if a node of the DT has no children, i.e., a leaf has been reached: the system suggests to the user to add the new concept as a child of that node, and if the user refuses, the procedure terminates.

As an aside, please notice that the algorithm is not a Depth First Search in a strict sense, since no procedure is implemented to backtrack towards the root of the DT if a leaf is reached and the user is not happy to add the new topic there.

\subsubsection{Entity Type Method} 
\label{entity-type-insertion}
This method is based on the idea of exploiting the Named Entity Recognition service provided by CNL: the Entity type that has been associated by CNL to the new concept is used to pick the best starting class of the Ontology and the corresponding node of the DT. For instance, if the concept ``American coffee" has been detected and its Entity type is \texttt{Food \& Beverages}, it would look unreasonable to start searching for the best position in the DT starting from the root: on the opposite, after properly mapping CNL Entity Types to corresponding classes in the Ontology, the algorithm can find a more convenient point to start Depth Search to find the most suitable parent node. Please notice that this mapping is performed offline for the whole Ontology through an automatic procedure not described here.
        
The method works as follows:
\begin{itemize}
    \item If the Entity Type is mapped to one of the classes of the Ontology, the system prompts the user if he/she agrees that the concept belongs to that class (e.g., ``Is \textit{American coffee} a kind of \textit{food or beverage}"). If the answer is positive, the research of the proper insertion point will start from the corresponding topic and not from the root of the DT;
    \item If the Entity Type is not mapped to one of the classes of the Ontology or it is mapped but the user does not agree that the concept belongs to that class, the method will simply call the Depth First method starting from the root. 
\end{itemize}

Notice that, even in the case that an appropriate starting node is found, a local search using the Depth First method will still be required to determine the most specific parent class: that is, after agreeing that ``American coffee" is a child of ``Food or Beverage", the system still needs to descend the DT to confirm that it is a kind of ``Beverage" and then a kind of ``Coffee".


When the correct insertion point is found, the actual addition procedure is started; if no insertion point is found, the procedure terminates without any insertion.
        
\subsubsection{Definitions and Synonyms Method} 
\label{definitions}
This method is based on the idea of asking the user for a definition of the new concept to be added, possibly using only one word, and to use the additional information to pick the best starting class of the Ontology and the corresponding node of the DT. 

To clarify how the method works, let's suppose once again that the new concept ``orange juice" has been extracted from the user sentence ``I love to drink \textit{orange juice} in the morning". The system will ask him/her ``I'm not sure what you are talking about. Please, try to define \textit{orange juice} with one word".

\begin{itemize}
    \item If the user is able to define the concept, e.g., by saying ``It is a kind of \textit{juice}" or ``Orange juice is a \textit{juice}", the sentence is sent once again to Dialogflow: a dedicated Intent (i.e., different from the previous ones) has been trained with the specific objective to extract definitions like this. In this case, \textit{juice} would be extracted. Once the definition has been extracted, first it is added on top of a stack, referred to as \textit{definition hierarchy} that keeps track of all the definitions given by the user related to this concept. Then, a check is performed to see if the new definition (or one of its synonyms retrieved from a Dictionary) corresponds to a class already in the Ontology: 
    \begin{itemize}
        \item If a matching class is found, the insertion point is assumed to be below that class: for instance, in case the user declared ``Orange juice is a kind of beverage", ``orange juice" shall be straightforwardly added as a child of ``beverage" since the latter already exists in the Ontology; 
        \item Otherwise, if no match is found (or no synonym corresponds to an already existing class), an additional definition is asked to climb the hierarchy upwards a more generic concept: since the user-defined ``orange juice" as a kind of ``juice" and such a class is not in the hierarchy, the user is asked again ``Please, try to define \textit{juice} with one word"; the question is re-iterated for each new definition by adding it on top of the \textit{definition hierarchy} until a matching class in the Ontology is found or the root is reached;
    \end{itemize}
    
    \item At any time, if the user doesn't know how to define a concept, sentences such as ``stop" or ``I don't know" are recognized by the system, and the Depth First Method is called starting from the root. 
    
    \item When a good starting point is found, it is still possible that a local Depth First search is required to find the most appropriate parent node. Notice however that, differently from other methods, we now have a full series of definitions that have been given by the user during the process and are contained in the definition hierarchy: i.e., ``Orange juice is a kind of juice" and ``Juice is a kind of beverage".  
    The addition procedure is repeated for each concept in the hierarchy in reverse order, from the most general to the most specific one, by taking them from the top of the stack. 
\end{itemize}
   
Thanks to the possibility of recursively defining all the required concepts not yet present in the Ontology, this is the only method that allows the addition of a set of new concepts at once: the result is a new branch in the DT, that will be attached below one of the already existing topics.
        
\subsubsection{Content Classification Method} 
\label{categories-method}
This method is based on the idea of exploiting the Content Classification service provided by CNL, plus the keyword mechanism used by the Ontology for Dialogue Management, Section \ref{sec:Ontology and DT}. The user is asked for a sentence that revolves around the new concept to be added: the method is different from the previous one since, in this case, the sentence is not necessarily a definition of the concept to be added, but can be anything that ``works well" with that concept. For instance, in case of ``orange juice", a sentence might sound like ``An orange a day keeps the doctor away!".
        
Once the user has provided the required sentence, the procedure queries CNL to obtain the sentence categories through the Content Classification service (as for the Entity Type method, categories have been mapped to corresponding classes in the Ontology and topics in the DT) and, in parallel, it searches for keywords in the sentence that match classes in the Ontology and topics in the DT. The following situations can occur: 

\begin{itemize}
    \item No category and no keywords associated with classes/topics are found;
    \item No category is found, but there are keywords associated with at least one class/topic in the Ontology/DT: if more than one matching class/topic is found, they are all returned, ordered from the highest to the lowest in the DT hierarchy; 
    \item At least a category is found, but no keywords are matching with any class/topic in the Ontology/DT: classes/topics with the strongest match, i.e., those with more categories in common with the sentence categories, are returned; if at least one is found, they are ordered from the highest to the lowest one in the DT hierarchy;
    \item At least a category is found, and additionally some classes/topics in the Ontology/DT match with the keywords contained in the sentence: the behaviour is the same as the previous point, but considering only classes/topics that match both with categories and keywords.
\end{itemize} 
        
Once the matching procedure has been executed:
\begin{itemize}
    \item All the classes/topics returned by the matching procedure are proposed to the user, one after the other, as a possible starting point: if the user agrees that the concept belongs to one of the proposed classes/topics, the local Depth First search will start from this class/topic and not from the root of the DT;
    \item If no match is found or the user does not agree that the concept belongs to any of the proposed classes/topics, the method will simply call the Depth First method starting from the root, as it has no clue on where the new concept could be inserted.
\end{itemize}  

This method has the advantage that the sentence provided by the user can later be stored in the Ontology and in the DT to be re-used by the system to talk about that concept/topic: either with the user itself or, after a mediator has checked and approved it, with other users possibly connected to the system.
 
\section{Experiments - Material and Methods}
\label{sec:experiments}
This Section presents the experiments conducted to assess the performance of the discussed solutions.

\subsection{Recognition of concepts}
\label{sec:experiment-DF}
The first experiment aims at evaluating the performances of the trained Dialogflow Agent when it comes to recognizing relevant concepts in the sentences provided as input. As already mentioned in Section \ref{sec:methodology-recognition}, the answers of 30 submitted questionnaires have been acquired through Amazon Mechanical Turk, for a total of $20\times 30$ sentences divided into the four classes ``Memories/Past", ``Preferences", ``Norms", and ``Beliefs". All sentences have all been transcribed and tagged: among those, the answers of the first 20 questionnaires have been used to train the agent, whereas the answers of the remaining 10 questionnaires have been set aside with the purpose of using them to test the Agent. The answers to the questionnaires used for training have been tagged by two independent researchers, while the answers of the remaining 10 questionnaires used for testing have been tagged by a third researcher. 


The Dialogflow agent has been trained using pieces of sentences, properly split using ``and" as a stop word or after 256 characters - the maximum number of characters allowed as input to Dialogflow: from now on, such pieces of sentence will be referred to as \textit{atomic sentences}. Splitting the sentence into atomic sentences is also expected to increase the probability of extracting at least one meaningful concept from the whole answer, which is what matters to expand the knowledge base (atomic sentences composed of single words or proper names are not tagged). 


To assess the performance of the Agent, the test sentences have been split according to the same rules used for the training sentences. Then, each atomic sentence in the test set has been fed to Dialogflow and outcomes have been collected. 

It shall be noticed that, whenever at least a concept is extracted from an atomic sentence, a procedure is started (not discussed in this article) to confirm that the user wants to add that concept, which will ultimately contribute to identifying relevant concepts even in presence of False Positives. Then, to evaluate DialogFlow response as a True/False Positive/Negative when processing an atomic sentence, we do not count individual concepts extracted or not extracted, but rather if the confirmation procedure will correctly/incorrectly start/not start for that atomic sentence. More specifically, for every atomic sentence, we say we have: 

\begin{itemize}
    \item a True Positive if the confirmation procedure correctly starts, i.e., if Dialogflow recognizes at least one concept that was actually tagged;
    \item a False Positive if the confirmation procedure incorrectly starts, i.e., if Dialogflow recognizes something but nothing was actually tagged;
    \item a False Negative if the confirmation procedure incorrectly does not start, i.e., Dialogflow does not recognize anything but something was tagged;
    \item a True Negative if the confirmation procedure correctly does not start, i.e., if Dialogflow does not recognize anything and nothing was tagged;
\end{itemize}


The rationale for the aforementioned classification is the following. 

Finding one or more TPs in a sentence is our primary objective as it means that Dialogflow correctly recognized at least one tagged concept that can be used to expand the Ontology. If at least one concept is correctly extracted, the atomic sentence can be safely labeled as TP because, through the subsequent interactive phase requiring user's confirmations, it is always possible to capture the relevant concepts even when both relevant and non-relevant concepts have been extracted by Dialogflow.

Suppose now that the system was not able to correctly extract any concepts among those that have been tagged. If there are some non-tagged concepts that have been extracted, then the atomic sentence is labeled as an FP: the confirmation procedure will start, incurring the risk of bothering the user with questions about irrelevant concepts he/she is not really motivated to insert. 

If no concepts have been extracted at all, but there are indeed some tagged concepts in that sentence, this shall be definitely considered as an FN. The confirmation phase will not start: we are losing the opportunity to expand the Ontology with relevant concepts, however, this is not too critical from the application perspective of long-term human-robot interaction. The user will never realize the missed opportunity, and will definitely be much less upset than having a system repeatedly asking questions to add irrelevant concepts.
 


Finally, if nothing was tagged and coherently nothing is detected, the atomic sentence can be safely labeled as TN.

%
%

The results of this experiment are presented and discussed in Section \ref{sec:results-recognition}.

\subsection{Insertion of concepts}
This test aims to compare the different insertion methods developed and evaluate their efficiency in terms of the number of interactions needed to insert a concept in the desired place. 

To test the four insertion methods independently from the concepts extracted by Dialogflow, for this experiment the concepts to be inserted into the Ontology have been taken from the \textit{Top 60,000 ``lemmas"} dataset, ordered by descending frequency, of Word Frequency Data\footnote{\href{https://www.wordfrequency.info/samples.asp}{https:\/\/www.wordfrequency.info\/samples.asp}}. For our purpose, the free version of the available datasets was satisfactory, as it is at one's disposal both in Excel (XLSX) and text (TXT) formats. The XLSX version allows filtering the lemmas by PoS (Part-of-Speech), obtaining only the nouns.
From the filtered nouns we kept the first 20 nouns having \texttt{OTHER} as Entity Type recognized by Cloud Natural Language, and the first 20 nouns having one of the mapped Entity Types, for a total of 40 most frequent nouns. The reason for this choice is that we want to assess whether there is some significant performance difference of the insertion methods if a concept is assigned a well-defined Entity type or only a generic one.

Insertion Methods 3 (Definitions and Synonyms) and 4 (Content Classification) require the user to provide a definition of the concept to be inserted or a sentence that revolves around it. To provide this information, a third person, not directly involved in research, is required. Due to the important role played by sentences in correctly identifying the insertion point, and the fact that we are now evaluating the performance of our solution from a technological perspective and not in terms of user experience, we decided to involve an expert for this task, instead of recruiting multiple naive participants. The selected person is a native English speaker, she has a high-school diploma in Classical studies, a Degree and a long professional experience in technical-scientific translation and finally a PhD in Bioethics, thus having the proved skills and sensitivity for this role. 

During the experiments, each concept has been inserted in the Ontology with each of the four developed methods. For all the concepts we kept track of the number of steps needed to add them below a previously established class: to the best of our knowledge, based on complete awareness of the Ontology structure, the chosen \textit{parent class} for each of the nouns to be inserted was the most appropriate one.

When the insertion is performed with the Depth First Method 1 starting from the root of the DT (Section \ref{brute-force}), the only information needed is the concept that should be inserted and the class under which it should be placed. As already mentioned, with this method the system starts asking ``yes/no" questions, and answers are provided by a researcher that knows the Ontology structure. The steps are counted starting from the first question asked (i.e., ``Is it correct to say that $<$concept$>$ is a type of object?"), to the confirmation of the insertion under the desired class (excluded). 


%
%
%

When exploiting the Entity Type Method 2 (Section \ref{entity-type-insertion}), the Entity Type associated with the concept is used to perform the research of the insertion point starting from the (hopefully) most convenient mapped class. Whenever the concept does not have any associated Entity Type recognized by CNL, the insertion method switches to a Depth First search starting from the root of the DT. In both cases, the steps are counted as already described above.

To perform the insertion with the Definitions and Synonyms Method 3 (Section \ref{definitions}), it is necessary to provide a definition of the concept to be added. Since researchers are necessarily aware of the target parent class, the aforementioned native English speaker expert has been asked to provide an objective definition of each concept without knowing how the Ontology is structured. The expert is asked to define each concept with one word (e.g., ``\textit{man} is a \textit{person}" or ``it is a \textit{person}"): then if no correspondences are found, she is asked for additional definitions until a correspondence is finally found and a whole branch of concepts/definitions is possibly inserted in the Ontology.
If the expert does not know how to define something, by saying a sentence such as ``stop" or ``I don't know" the insertion procedure switches to the Depth First search starting from the root of the DT. 
Since this method, differently from previous ones, may insert several concepts at once when the insertion procedure terminates, the average number of steps per concept in the branch is reported (instead of the total number of steps for just one concept) to allow for a fair comparison with other methods. 

As the previous method, the Content Classification Method 4 (Section \ref{categories-method}) requires the user to provide a sentence revolving around the concept that should be inserted. Such a sentence is then sent to CNL to extract its category (if present), which is used to find the most appropriate starting class in the Ontology. Again, since researchers are aware of the destination classes, the Ontology structure, and how CNL extracts the categories from the sentences, the expert has been involved to formulate sentences as needed.
Whenever the sentence does not have an associated category or no mapping with the Ontology is found, the insertion method switches to a Depth First search starting from the root of the DT. In both cases, steps are counted as in Methods 1 and 2.

The results of this experiment are presented and discussed in Section \ref{sec:results-insertion}.

\section{Experiments - Results and Discussion}
\label{sec:results}
This section presents and analyzes results.

\subsection{Recognition of concepts}
\label{sec:results-recognition}
\begin{figure}
    \centering
    \includegraphics[width=\linewidth]{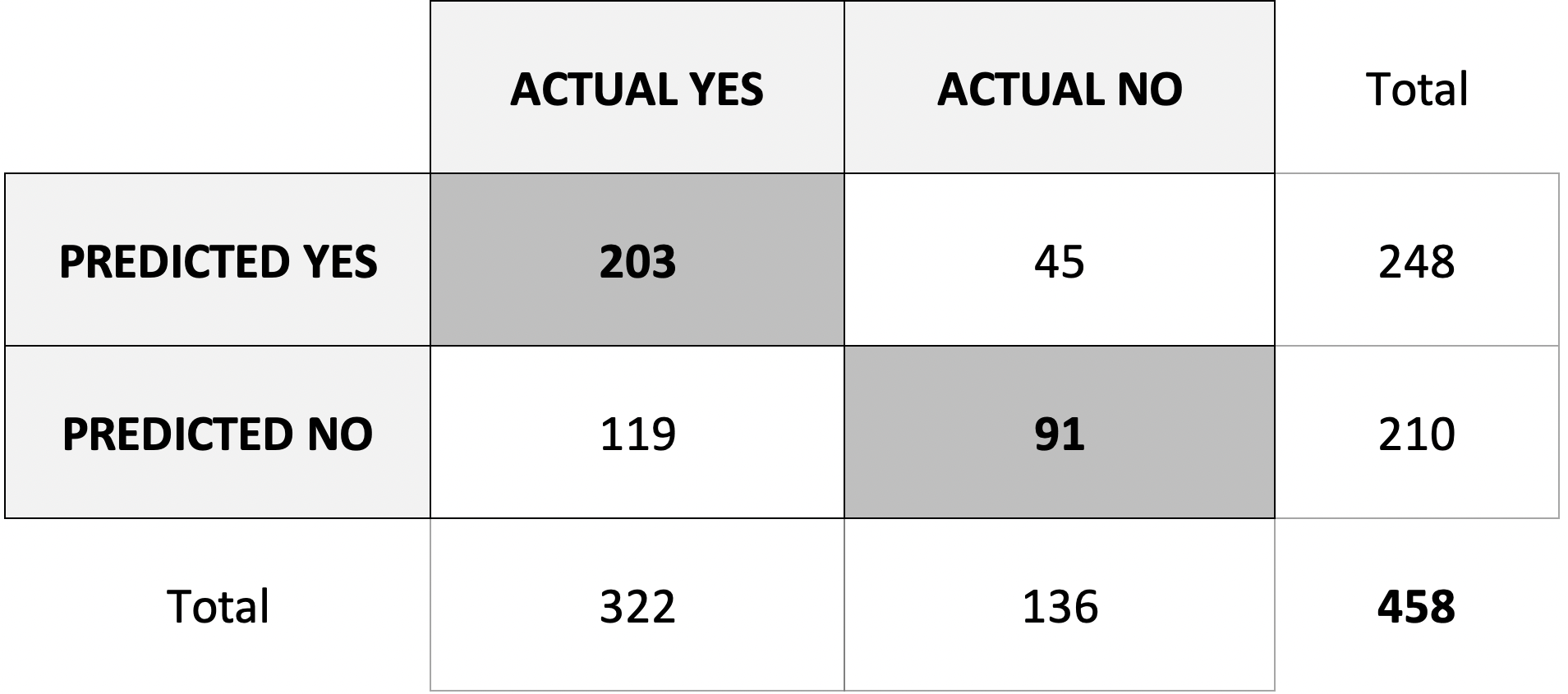}
    \caption{Confusion Matrix summarizing the results of classification of atomic sentences.}
    \label{fig:confusion-pieces}
\end{figure}

The results of the classification of atomic sentences performed by Dialogflow are summarized in the Confusion Matrix shown in Figure \ref{fig:confusion-pieces}. The cell \textit{Actual yes/Predicted yes} reports the number of $TP$ (i.e., True Positives), meaning that the confirmation procedure would correctly start since at least one tagged concept in the atomic sentence has been correctly detected. The cell \textit{Actual no/Predicted yes} reports the number of $FP$ (i.e., False Positives), meaning that the confirmation procedure would start but no tagged concept has been correctly extracted. The cell \textit{Actual yes/Predicted no} reports the number of $FN$ (i.e., False Negatives), meaning that the confirmation procedure will not start even if at least one concept was tagged in the atomic sentence. The cell \textit{Actual no/Predicted no} reports the number of $TN$ (i.e., True Negatives), meaning that nothing was tagged and, coherently, the confirmation procedure will not start.

It shall be noticed that the vocal answers collected from recruited workers of Amazon Turk are very different in their length and syntactic structure since we let participants free to answer questions in the way they prefer, with no constraints. As an example, consider two replies to the question \textit{Please tell me about your childhood} in the Memories/Past Intent (sentences are split into atomic ones using the ``and" stop word).

The first example is composed of two atomic sentences.

\begin{enumerate}
    \item \textit{My childhood} was great I grew up with \textit{three other siblings} I was the second oldest sibling and...
    \item I grew up \textit{\textbf{playing soccer}} and... 
    \item tennis
\end{enumerate}


Tagged concepts are in Italics and extracted concepts are in bold. In this case, the concepts ``My childhood" and ``three other siblings" were tagged in the first atomic sentence but were not extracted, whereas the concept ``playing soccer" was tagged and correctly extracted in the second atomic sentence. The first atomic sentence is labeled as FN, the second one as TP, the third one as TN. 


A more complex example composed of five atomic sentences is the following.

\begin{enumerate}
    \item I have quite a mixed childhood I was born in \textbf{\textit{Belgium}} I grew up in \textit{Norway} until I was about 9 years old and... 
    \item then I moved to \textit{the UK} I was very happy in my \textbf{early childhood I moved schools} a lot but that meant that I have \textit{lots of friends} in different places which was really fun and...
    \item then as I became a teenager my parents split up and...
    \item things became a lot more difficult and... 
    \item it was a lot \textit{more troubled time} for me.
\end{enumerate}

In this case, the first atomic sentence is a TP because at least one tagged concept has been correctly extracted; the second atomic sentence is labelled as an FP because the non-tagged concept ``early childhood I moved schools" has been extracted and this will inappropriately start the confirmation procedure; the third and fourth sentences are TN, and finally the fifth sentence is an FN. 



\subsubsection{Data analysis}
\label{Data analysis}

Starting from the Confusion Matrix, the most common parameters to evaluate the performance of binary classifiers have been computed and are reported in the following. 



\textit{Accuracy}, which is computed as $(TP+TN)/(TP+TN+FP+FN)=0.64$, and indicates the ratio of times that the confirmation procedure starts \textit{if and only if} it is needed, is not very high: this is due to the considerable number of atomic sentences classified as FN. The high number of FN negatively affects also the \textit{Sensitivity}, which is computed as $TP/(TP+FN)=0.63$ and indicates how often the confirmation procedure correctly starts when there is at least one tagged concept. 

The value of the \textit{Specificity}, computed as $TN/(TN+FP)=0.67$, indicates the ratio of times that the confirmation procedure does not start when nothing is tagged: its value is negatively affected by a considerable number of FPs. 
\textit{Precision}, computed as $TP/(TP+FP)=0.82$, is good enough: it indicates that almost every time the confirmation procedure starts, there is at least a tagged concept that has been correctly extracted.

The Matthews correlation coefficient, which is generally regarded as the best way of describing the confusion matrix by a single number, even in presence of unbalanced samples, turns out to be $MCC=0.27$ in a range between -1 and +1, where +1 indicates perfect prediction (the confirmation procedure starts exactly when needed), 0 no better than random prediction and -1 total disagreement between prediction and observation. It has also be argued that, when MCC equals neither -1, 0, or +1, it cannot be considered as a reliable indicator to measure the similarity of the predictor to random guessing. In our case, MCC shows that predictions are not random while being far from having optimal performance. 


To summarizing, results are not as good as we might desire: this is likely to depend on the enormous variety that can be observed in respondents' replies, which makes it difficult for Dialogflow to find common syntactic patterns to extract relevant concepts. For a similar reason, during the training phase, it was not possible to give the two independent taggers precise instructions on how to select relevant concepts to be extracted: then, taggers were necessarily asked to operate based on the structure of the sentence and their ``feeling" of what was worth being emphasized, which added a significant amount of fuzziness to the whole process. 

Some additional considerations shall be made in light of the specific application for which the whole system is designed. 

First, it can be observed that performance is mostly affected by the high number of FNs. Generally speaking, having a low number of FNs is highly desirable and critical in many applications (think about a test for Covid-19 that repeatedly fails to identify infected people). However, in our case, having some FNs in an atomic sentence may not be that critical: this corresponds to missing an opportunity for knowledge acquisition, but it does not have a dramatic impact on the system's performance, as the user might never realize the missed opportunity.

Concerning the number of FPs, it is straightforward to notice that keeping this number low is crucial to avoid bothering the user with too many questions. However, asking for a confirmation definitely helps to mitigate the negative effects of FPs, and the fact of engaging the user in a conversation may have a beneficial effect, in the case of a Socially Assistive Robot whose main purpose is improving mental health and reducing loneliness.

\subsection{Insertion of concepts}
\label{sec:results-insertion}

\begin{figure}[h!]
    \centering
    \includegraphics[width=\linewidth]{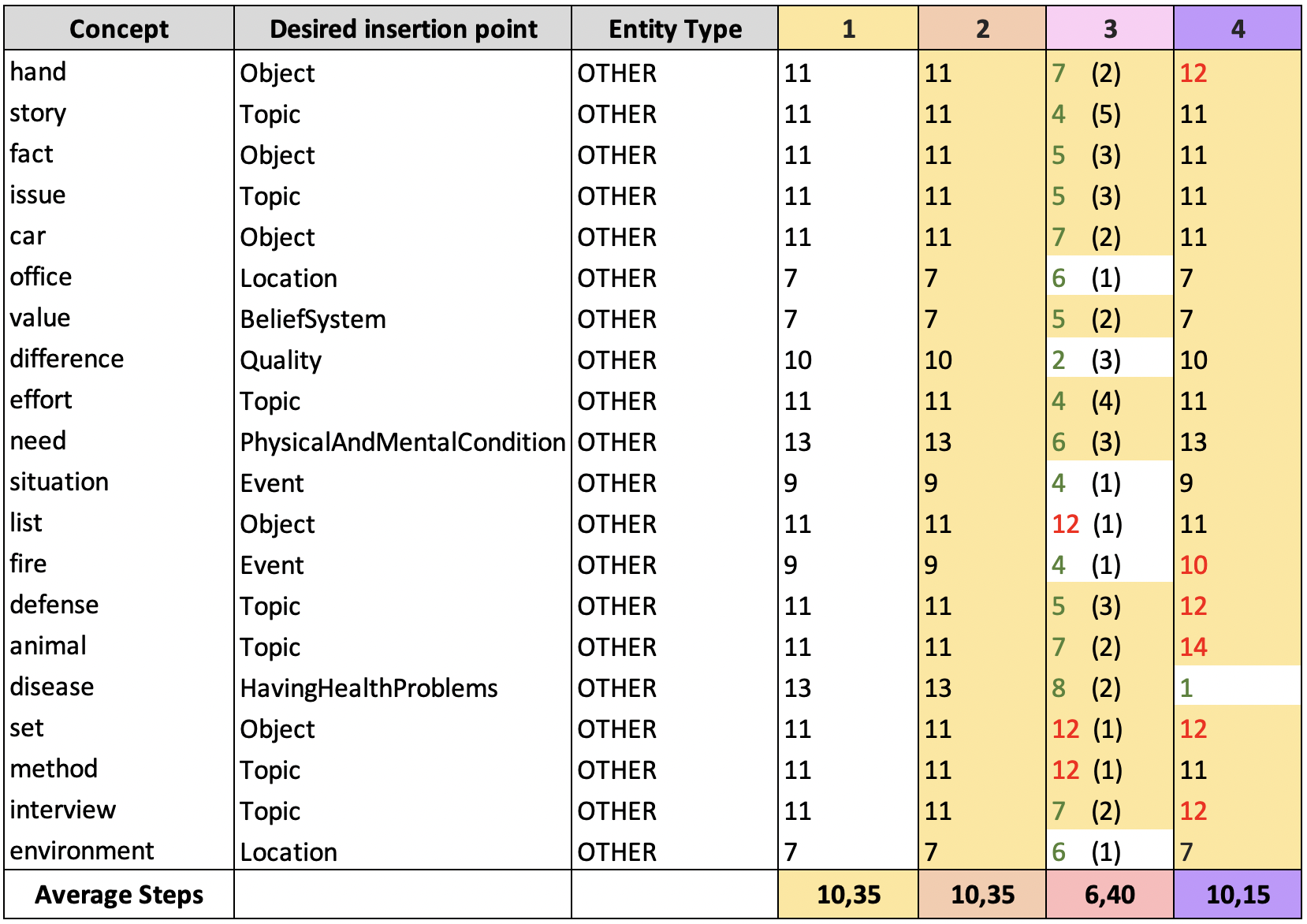}
    \caption{Number of steps required to insert the concepts when the Entity Type is OTHER.}
    \label{fig:steps-other}
\end{figure}

\begin{figure}[h!]
    \centering
    \includegraphics[width=\linewidth]{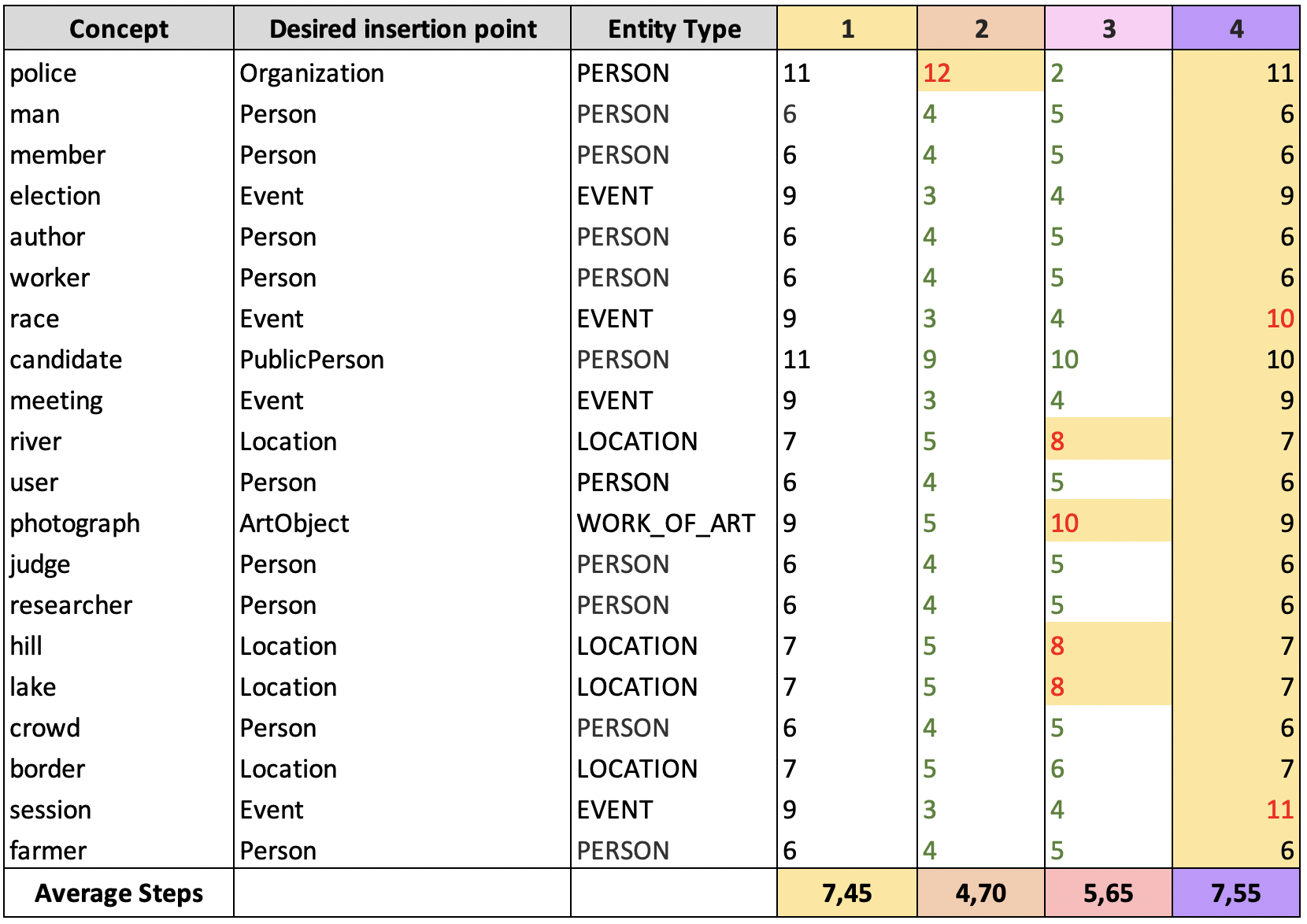}
    \caption{Number of steps required to insert the concepts when the Entity Type is a mapped one.}
    \label{fig:steps-mapped}
\end{figure}

The four insertion methods have been compared by computing the following quantities:

\begin{enumerate}
    \item The average number of steps required for the insertion of nouns with Entity Type \texttt{OTHER} according to CNL (Figure \ref{fig:steps-other});
    \item The average number of steps required for the insertion of nouns with mapped Entity Types according to CNL (Figure \ref{fig:steps-mapped}).
\end{enumerate}

Figure \ref{fig:steps-other} reports the first 20 nouns with Entity Type \texttt{OTHER}, while the table in Figure \ref{fig:steps-mapped} presents the first 20 nouns with a mapped Entity Type.

The first column of each table contains the nouns, the second column contains the class under which the noun should be inserted according to researchers that designed the Ontology (this class is known a priori only when using Method 1 and 2), the third column contains the Entity Type (exploited only by Method 2), while the last four columns contain the number of steps required for the insertion corresponding to each of the four methods. 
The last row reports the average number of steps required by each method. 

%
%
%


The Insertion Method 1 is simply a Depth First search starting from the root of the DT: it will be taken as a baseline for comparison and therefore does not require additional comments. 

When exploiting the Entity Type Method 2, 
if the concept to be inserted has an Entity Type mapped to a class in the Ontology (third column in Figures \ref{fig:steps-other} and \ref{fig:steps-mapped}), this information is used to start the research of an insertion point from that class using a local Depth First search, and therefore the search itself is expected to be shorter. 

However, it is still possible that a Depth First search starting from the root of the DT is needed (i.e., switching to Insertion Method 1, yellow cells). This happens in two situations:
\begin{itemize}
    \item The Entity Type is \texttt{OTHER} (concepts in Figure \ref{fig:steps-other}): there is no useful information contained in this type to ease the research of the starting point. In this case, the number of steps will be the same as the insertion Method 1;
    \item The Entity Type is different from \texttt{OTHER} (concepts in Figure \ref{fig:steps-mapped}) but it is not coherent with the desired starting class (second column). An example of this eventuality is represented in the first row: the desired starting class of the concept ``police" is ``Organization"; however, CNL labels ``police" as \texttt{PERSON}: the first question asked to the user will be: ``Is it correct to say that \textit{police} is a type of \textit{person}?". The answer of the user will be ``no", and a Depth First search will be called from the root to give the user the possibility of inserting the concept where desired. Unfortunately, this inconsistency increases by one the number of steps needed: in the case of ``police" the steps become 12 instead of 11, and the number is written in red to indicate a worse performance with respect to Insertion Method 1. 
\end{itemize}

The cells that are not coloured indicate that a good starting class for insertion has been found. In all these cases, the number of steps required to insert the concept under the desired class is lower with respect to Method 1. 
Results corresponding to Method 3 are reported in the last but one column of the tables in Figures \ref{fig:steps-other} and \ref{fig:steps-mapped}. In this case, it is also reported the number of inserted concepts (in round brackets): this information is fundamental as it distinguishes this method from the others as it gives the possibility of adding multiple concepts at once in the Ontology, creating a new branch in the DT. 
The number of steps required for the insertion using this method is consequently computed as the total steps divided by the number of inserted concepts to allow for a fair comparison.

Also in this case there is a visible difference between nouns, depending on their Entity Type: the nouns with Entity Type \texttt{OTHER} reveal to be more difficult to be defined with respect to the nouns with a mapped Entity. Most times, in the first group, multiple definitions were necessary to find a good starting point for the local Depth First search. In the second group, almost every time, the first definition given by the person coincided with the Entity Type provided by CNL, and therefore the number of steps is the same as the number of steps needed with Method 2 plus one: the first definition asked.

Again, yellow cells indicate that a Depth First search starting from the root of the DT has been used, and a red/green number means that the insertion required more/fewer steps with respect to Method 1. Nouns with Entity Type \texttt{OTHER} require a Depth First search starting from the root significantly more frequently than nouns with a mapped Entity Type: this means that, at some point, the person did not know which definition to provide. For the second group of nouns, as we already said, one definition was almost always enough to find a good starting class. The only exceptions are the nouns ``river", ``photograph", ``hill" and ``lake": the person defined these concepts as \textit{things}, hence the Depth First search starting from the root was triggered and the overall insertion required one additional step with respect to Insertion Method 1.

Results corresponding to Method 4 are reported in the last column of the tables in Figures \ref{fig:steps-other} and \ref{fig:steps-mapped}. Again, the yellow cells mean that a Depth First search starting from the root has been used for the insertion: either no results from the Content Classification service were returned, or the result returned did not correspond to a correct class under which the noun could be placed. Only in one case, the sentence was classified with the same category of the correct class: the sentence ``Infectious diseases can sometimes be deadly" related to the concept ``disease" was classified as belonging to \texttt{Health/Health Conditions/Infectious}, which is the same classification of the Ontology class \textit{HavingHealthProblems}. 

As a Depth First search starting from the root has been used for all the insertions (except one), and in some cases, additional steps were performed due to wrong matching classes, Method 4 is definitely worse than Method 1. 

\subsubsection{Data analysis}
\label{Data analysis}
In order to perform statistical data analysis, the normality of each dataset has been tested with the Shapiro-Wilk Test: none of the samples was normally distributed. Hence, to assess significant differences between the four methods, the Wilcoxon signed-rank test has been used: this test is an alternative to the paired t-test that makes no assumptions on the distribution of the data. The null hypothesis to be rejected with the Wilcoxon signed-rank test asserts that the medians of the two samples are identical. 

For each test performed, both a \textit{W-value} and \textit{z-value} have been computed to pairwise compare insertion methods. The \textit{z-value} (or \textit{z-score}) is then converted into a \textit{p-value}, which is the probability to make an error by rejecting the null hypothesis (in other words, a lower p-value means that there is a more significant difference between the samples). We aim to reject the null hypothesis with $p \leq 0.01$. 

\begin{figure}[h!]
    \centering
    \includegraphics[width=\linewidth]{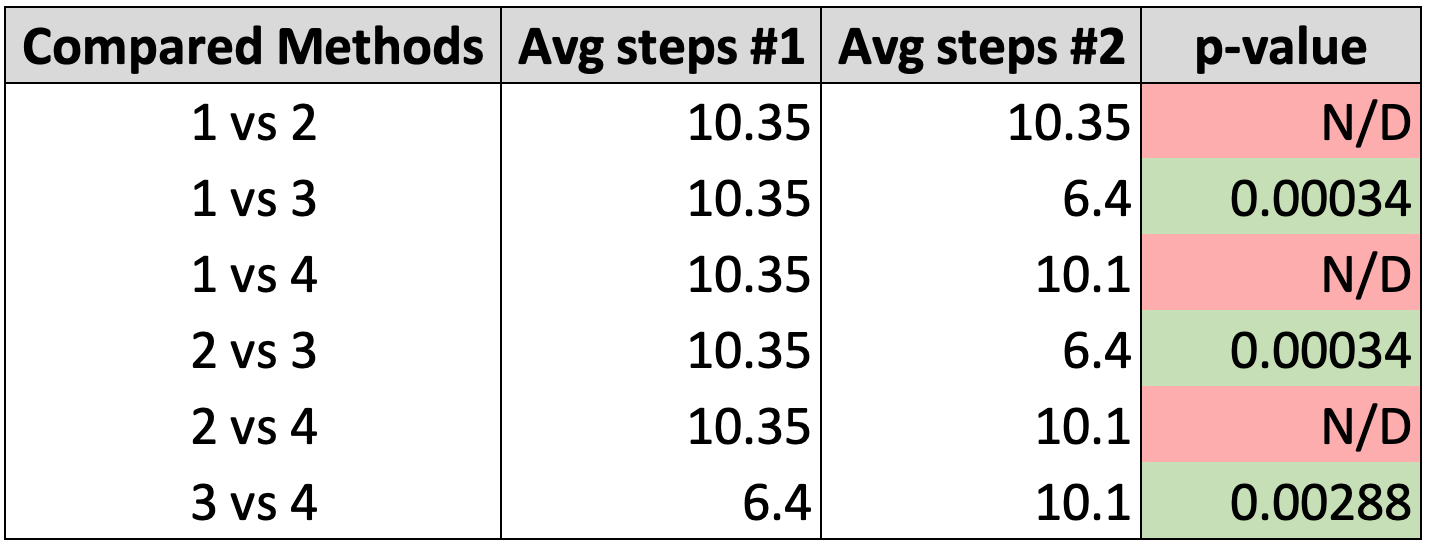}
    \caption{Results of the Wilcoxon signed-rank test when comparing Insertion Methods for nouns with Entity Type \texttt{OTHER}.}
    \label{fig:comparison-no-entities}
\end{figure}

First, we consider the insertion of the nouns with Entity Type \texttt{OTHER} not mapped to an Ontology class, Figure \ref{fig:steps-other}.
The Table in Figure \ref{fig:comparison-no-entities} shows the results of the pairwise comparisons between the four insertion methods. 

The Wilcoxon signed-rank test can neither be used to compare Method 1 and Method 2 nor to compare Method 1 with Method 4: these samples contain too many identical values (ties). When data include tied values, they are ignored and the calculation of the Wilcoxon signed-rank statistic is made on the remaining values. Method 1 and Method 2, in this case, have the same number of steps, since Method 2 immediately switches to Method 1 with nouns having Entity Type \texttt{OTHER}. Similarly, also Method 1 and Method 4 contain many tied values: not considering ties brings the size of the samples to 7, which is not large enough for the distribution of the Wilcoxon W statistic to follow a normal distribution and, more in general, to provide reliable results. 


The glaring result obtained with this test is that Insertion Method 3 is significantly better than all the other methods when it comes to inserting concepts without a mapped Entity Type: the definition provided by the user makes up for the absence of a mapped Entity Type and plays a key role to find a good starting class in the Ontology for the insertion.

\begin{figure}[h!]
    \centering
    \includegraphics[width=\linewidth]{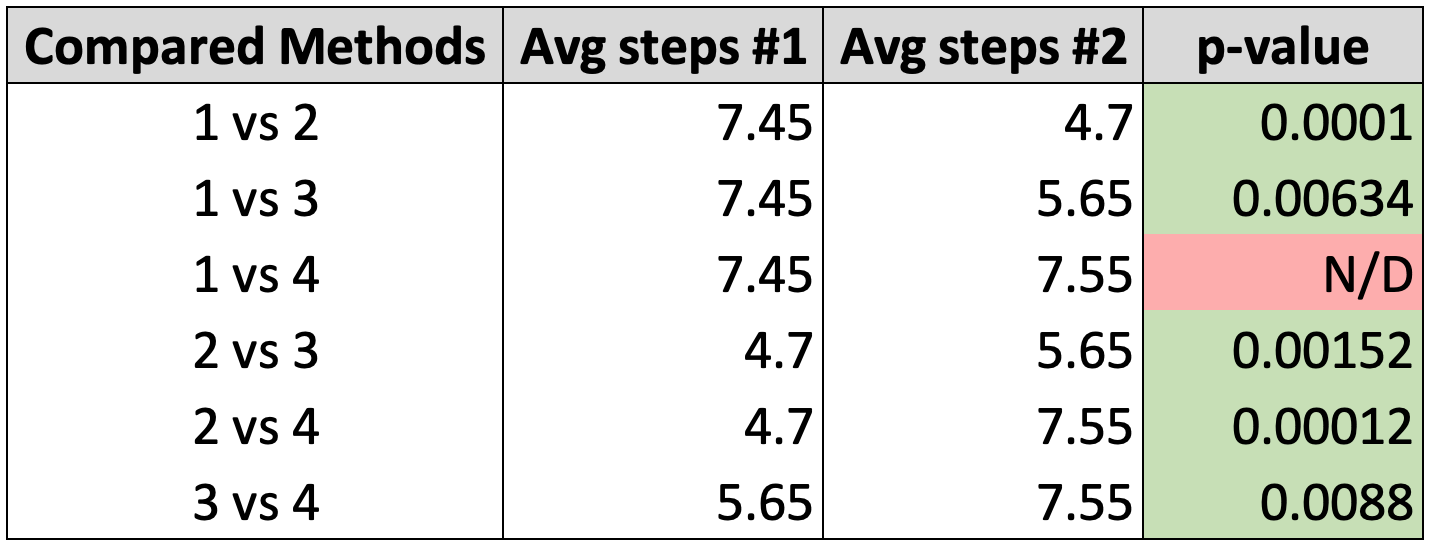}
    \caption{Results of the Wilcoxon signed-rank test when comparing Insertion Methods for nouns with a mapped Entity Type.}
    \label{fig:comparison-entities}
\end{figure}

Second, we consider the insertion of the nouns with mapped Entity Type, already shown in Figure \ref{fig:steps-mapped}.
The table in Figure \ref{fig:comparison-entities} shows the results of pairwise comparisons between the four insertion methods. 

Starting from the first row, the extremely low p-value indicates that there is a significant difference between Method 1 and Method 2. As it could be expected, Method 2 performs significantly better when it comes to inserting concepts with a mapped Entity Type. When comparing Method 1 with Method 3, again there is a significant difference (in favour of Method 3), motivated by the fact that these nouns are easier to define, and almost always the first definition provided allowed to find a good starting class in the Ontology.

Methods 1 and 4 could not be compared because data included many tied values (all but 3), hence the sample size was too small to allow a reliable calculation of the Wilcoxon signed-rank statistic.

Method 2 turns out to perform significantly better than 3 and 4, and Method 3 turns out to perform significantly better than 4.


\section{Conclusions}
\label{sec:conclusion}
The article proposes an approach to (i) recognize and extract meaningful concepts from the users' sentences during the interaction with the system, as well as (ii) four methods allowing the run-time insertion of the recognized concepts into the knowledge base. The solutions proposed, partly based on the third-party services Dialogflow and Cloud Natural Language, have been integrated into the system for culture-aware interaction that has been developed in the course of the CARESSES project for assisting older people in domestic and care home environments.

Experiments have been performed to assess the performance of the method for concept extraction as well as the four methods developed to insert the recognized concepts in the Ontology in run-time. 

The results of the first set of experiments demonstrated that the system achieves a sufficiently high Precision, which means that recognized concepts usually correspond to concepts that would also be extracted by a human expert. On the other side, Accuracy, Sensitivity, Specificity, and the Matthews correlation coefficient are far from being optimal. Considering the enormous variety in the sentences acquired from Amazon Mechanical Turk workers for training and testing the system, and the consequent complexity of properly identifying and tagging concepts to be extracted, this is not a surprise. However, the negative impact of the high number of False Negatives and False Positives is partially mitigated by the application for which the system is designed. FNs have not a dramatic effect as they simply correspond to missed opportunities for acquiring knowledge from the user. 
Similarly, FPs are not that critical because the user is always asked to confirm if he/she really wants to add the extracted concept to the knowledge base (even if, generally speaking, it is important to keep their number low to avoid bothering the user with too many questions).


The results of the second experiment have shown that, when the concept to be inserted has a well-specified Entity Type according to third-party software for Natural Language Processing (i.e., Google Cloud Natural Language), this may significantly help to insert the concept in the right place of the Ontology; if not, multiple definitions provided by the users can be fed to a Dialogflow Agent to achieve similar performance, with the additional benefit of adding more concepts to the knowledge base to be re-used in subsequent interactions. 

The article does not present experiments with recruited participants to investigate how the process can be expanded to multiple users for knowledge crowdsourcing, nor subjectively evaluates the user's experience while interacting with the system: both areas are currently subject of research.

\vspace{10mm}
\noindent \textbf{Conflicts of Interest}
The authors declare that there is no conflict of interest.

\end{document}